\pdfoutput=1

\documentclass[11pt]{article}

\usepackage{acl}

\usepackage{times}
\usepackage{latexsym}

\usepackage[T1]{fontenc}

\usepackage[utf8]{inputenc}

\usepackage{microtype}

\usepackage{inconsolata}

\usepackage{multirow,multicol}
\usepackage{tablefootnote}
\usepackage{siunitx}
\usepackage{booktabs}
\usepackage{graphicx}
\usepackage{subfigure}

\title{Efficient Language Modeling for Low-Resource Settings \\ with Hybrid RNN--Transformer Architectures}

\author{Gabriel Lindenmaier \\
  \texttt{gabriel.lindenmaier@proton.me} \\ \\
  \bf
  Sean Papay \\
  Fundamentals of Natural Language Processing \\
  University of Bamberg \\
  Germany \\
  \texttt{sean.papay@uni-bamberg.de} \\ \\
  \bf
  Sebastian Padó \\
  Theoretical Computational Linguistics \\
  University of Stuttgart \\
  Germany \\
  \texttt{sebastian.pado@ims.uni-stuttgart.de} \\}

\begin{document}
\maketitle
\begin{abstract}
  Transformer-based language models have recently been at the forefront
  of active research in text generation. However, these models' advances
  come at the price of prohibitive training costs, with parameter counts in the
  billions and compute requirements measured in petaflop/s-decades.
  In this paper, we investigate transformer-based architectures for
  improving model performance in a low-data regime by selectively
  replacing attention layers with feed-forward and quasi-recurrent
  neural network
  layers. We test these
  architectures on the standard Enwik8 and Wikitext-103 corpora.
  Our results show that our reduced architectures outperform
  existing models with a comparable number of parameters, and
  obtain comparable performance to larger models while significantly 
  reducing the number of parameters. The implementation is going to be 
  available at this URL\footnote{\url{https://github.com/lindenmg/Hybrid-RNN-Transformer}}
\end{abstract}


\section{Introduction}
\label{sec:introduction}

The advent of deep neural networks, and in particular the
transformer architecture \citep{transformer}, has led to significant
advancements in text generation areas such as conversation agents
\cite{Cohen2022Lambda}, commercial key-word based text-generation
\cite{Shao2021Ecommerce}, news generation \cite{Cao2019News} and
machine translation \cite{bahdanau15:_neural_machin_trans}. And since late 2022 general purpose Large Language Models (LLM) like LLama 3.3 \cite{grattafiori2024llama3herdmodels} and Qwen 2.5 \cite{qwen2025qwen25technicalreport}, which generate answers based on natural language prompts \cite{ouyang2022training, schulhoff2024promptreportsystematicsurvey}. A central
building block of these systems are neural language models (NLMs) with
an ever increasing number of parameters. This leads to a scaling
bottleneck: to obtain a linear increase in performance, the number of
parameters, training data size and computation costs increase exponentially
\citep{kaplan2020scaling,Rae2021Scaling}.
Solutions for this scaling problem are critical improving
the environmental impact of such models \citep{10.1145/3381831}.

In this paper, we investigate the smarter combination of existing
architectural building blocks: while the attention mechanism of the
transformer enables every token to attend to every other token, more
traditional recurrent networks (RNNs) have a temporal inductive bias
on account of their sequential nature. This makes it easier for them to
learn the underlying structures of natural language, at the cost of a
more limited context memory. Our hypothesis is that RNNs are sufficient for
\textit{initial} layers of neural networks models, which learn
comparatively simple structures like part-of-speech tags
\cite{tenney-etal-2019-bert}, while the more powerful transformers are
more useful for later, ``deeper'' stages of the neural network.
Consequently, there is reason to believe that RNN--transformer mixture
architectures might be superior to pure transformers --- at least in
terms of effective use of parameters and training time.

This hypothesis is based on the Single-Headed-Attention LSTM (SHA-LSTM) from \citet{merity2019single}. The SHA-LSTM is a LSTM model, therefore a RNN-model, followed by a single attention head layer at the end. The SHA-LSTM shows comparative performance to transformers. Through our own attempts to reproduce the SHA-LSTM we concluded the following: the largest amount of RNN layers that still scales well in depth, followed by transformer blocks, should improve upon pure transformer baselines.

Concretely, we experiment with three models: Our transformer baseline is an adapted version from \emph{Pay Attention when Required} by \cite{mandava2020pay}. We call it PAR Transformer for simplicity. We use this architecture as a baseline because it is already more efficient in regards to its sequential computation. Building upon this baseline,
and motivated by our hypothesis about recurrent layers, we propose the
Hybrid Transformer architecture, which replaces part of the
transformer with two recurrent layers at the beginning
of the model. Finally, our own version of the Single-Headed-Attention LSTM, a multi-layer RNN with an additional attention head. We find that, in particular, the Hybrid
Transformer outperforms existing models with a comparable number of
parameters, and obtains comparable performance with a significant
decrease in parameters \cite{10.1145/3381831}.


\section{Related Work}
\label{sec:related_work}
To improve the efficiency of transformers, we combine self-attention
layers with RNN layers.  We now outline similar approaches from the
literature.

\citet{merity2019single} combines an LSTM with single-headed attention
blocks. The resulting Single-Headed-
Attention LSTM (SHA-LSTM) model is trainable using one GPU and
achieves respectable results at language modeling. Merity's main
objective is to accelerate the training convergence of language
models, given that transformers are too slow for one GPU.  Despite the
SHA-LSTM's improvements in efficiency, it is still not competitive
with the quantitative performance of transformers. The contemporary
Transformer-XL, for example, achieves a slightly better result with
fewer parameters \citep{dai-etal-2019-transformer}, although it takes
4 days with 4 GPUs to train \citep{lei2021attention}.

\citet{lei2021attention} combines single-headed attention with a fast RNN
variant called the Simple Recurrent Unit (SRU) and achieves SOTA
performance on the enwik8 and Wikitext-103 datasets. Like the QRNN,
the SRU by \citet{lei-etal-2018-simple} uses parallelizable gates. The
sequential section of this RNN variant also only applies vector
operations. In addition, it uses a highway connection along the temporal axis of one SRU
layer. Using a custom, scaled weight initialization,
it can accomodate far more layers than classical RNN architectures. The
SRU++ replaces the linear input projection of the SRU with a
single-headed attention adaption of \citet{transformer}. Like
\citet{merity2019single}, \citet{lei2021attention} finds that one
attention head at the last layer provides most of the performance
gain.

\citet{lei2021attention} choose their parameters for higher
comparability with Transformer-XL by
\citet{dai-etal-2019-transformer}.  They find that the SRU++
outperforms Transformer-XL in all tasks, while being 3-8 times faster
to train (although a full CUDA implementation is used).
\citet{Hutchins2022BlockRnnTf} propose the Block-Recurrent Transformer,
which combines RNNs with transformers. They use a
transformer layer as an RNN cell, taking input not token-, but
block-wise. This significantly outperforms the Transformer-XL from
\citet{dai-etal-2019-transformer} on long-range tasks with half the
computation requirements.

\citet{hao-etal-2019-modeling} propose the bi-ARN, which combines a bi-directional RNN with limited recurrent depth with a self-attention layer for Neural Machine Translation (NMT). But the RNN is used in parallel to the transformer encoder and as gradient skip connection. They do not control for number of parameters nor training time: Their best model shows a significant parameter increase.
\citet{chen-etal-2018-best} propose a stacked RNN-Transformer encoder for NMT, which needs a pretrained and frozen RNN part. Although Chen et al.’s base architecture of improvement, the RNMT+ needs 1.66x more training time than their transformer baseline.

\citet{so2021primer} conduct an automated architecture search on
transformer language models. The finding which improved
performance the most was squaring the ReLU activation in
feed-forward blocks. However, the second best improvement is more notable,
as they used masked depth-wise convolutions with width of three after
the Q, K \& V projections in the attention head. This means the
transformer had, for each token, direct information about the two
predecessor tokens.

\section{Methods}
\label{sec:architecture_datasets_main}

In this section, we describe the architecture of the three language
models we experiment with and the building blocks they consist
of.

\begin{figure*}[ht]
  \centering
  \subfigure[Feed-forward layer block]{\includegraphics[scale=0.75]{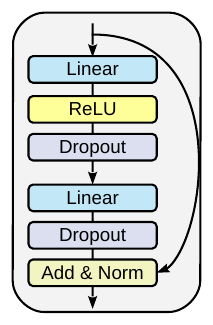}}
  \hfill
  \subfigure[Multi-head attention]{\includegraphics[width=0.4\textwidth]{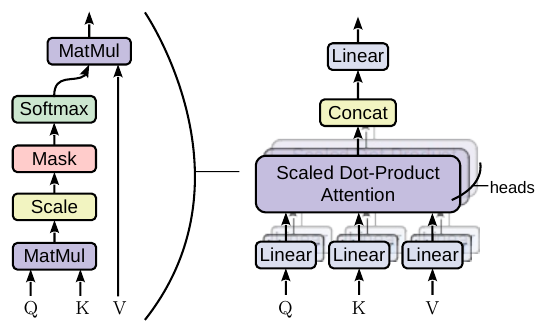}}
  \hfill
  \subfigure[QRNN cell]{\includegraphics[width=0.42\textwidth]{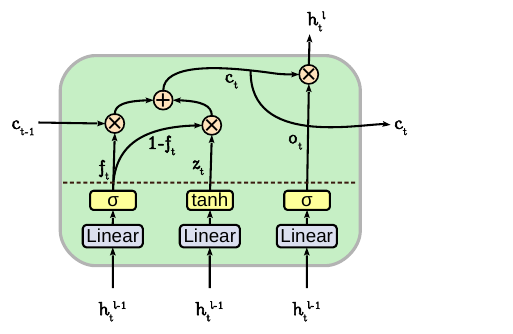}}
  \caption{Building blocks of our models}
  \label{fig:model_layers}
\end{figure*}

\paragraph{Building Blocks.} We compose our models out of four types
of elementary building blocks: a feed-forward layer, relative multi-head attention layer, recurrent QRNN layer and RNN-Dropout layer. We notate them as \texttt{f},
\texttt{a}, \texttt{q}, and \texttt{|}
(cf. Table~\ref{tab:architectures} and also Figure~\ref{fig:model_layers}).  

The first, \texttt{f}, is a
two-layer feed-forward network similar to that used in 
the transformer of \citet{transformer}.
The first layer (boom layer) scales the feature dimensionality up to
the hidden size, while the second layer scales it back down to the
original embedding size, obtaining a refined representation. We use the
fast ReLU activation \cite{Fukushima1975Relu} after the boom layer.
Dropout \cite{Srivastava2014Dropout} is applied after both layers, and 
a residual connection is used around the whole feed-forward block.
Layer-normalization \cite{Xu2019Layernorm} is applied to the
residual connection \cite{liu2020tftraining}.

Our second building block, \texttt{a}, is an multi-headed attention encoder layer \citep{dai-etal-2019-transformer} with access to relative position information, i.e.,~the positional difference between the current token and the attended one.
The attention layer also uses dropout after the attention and a residual connection with layer-normalization.

Our third building block, \texttt{q}, is a quasi-recurrent layer
\citep{bradbury2016quasi}, a partially parallelizable RNN variant.
Specifically, we use an AWD-QRNN, a quasi-LSTM with
weight dropping, although we do not use ASGD to train
\citep{merity2018regularizing}. This layer uses weight dropout based
on DropConnect \citep{10.5555/3042817.3043055}, which is applied to
the weights of the three gates.


Finally, \texttt{|} denotes an RNN-Dropout layer, which
drops fixed individual dimensions of the hidden embedding vector
sequence.



\begin{figure*}[h!t]
  \centering
  \subfigure[Base model layout]{\includegraphics[scale=1.0]{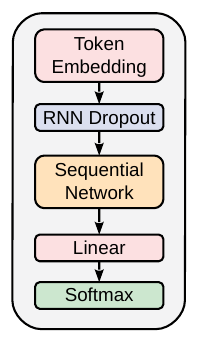}}
  \hfill
  \subfigure[PAR transformer block]{\includegraphics[scale=1.0]{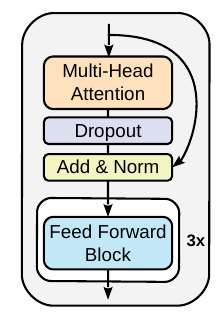}}
  \hfill
  \subfigure[QRNN- plus feed-forward layer (QRNN FF)]{\includegraphics[scale=1.0]{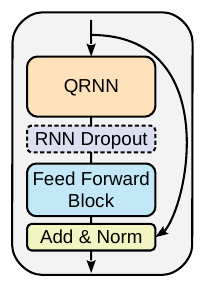}}
  \hfill
  \subfigure[QRNN block for the Hybrid Transformer]{\includegraphics[scale=1.0]{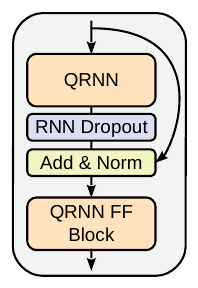}}
  \caption{Language model buildup \& single PAR transformer and QRNN blocks.}
  \label{fig:building_blocks}
\end{figure*}

\begin{figure*}[h!t]
  \centering
  \subfigure[Attention QRNN]{\includegraphics[scale=1.0]{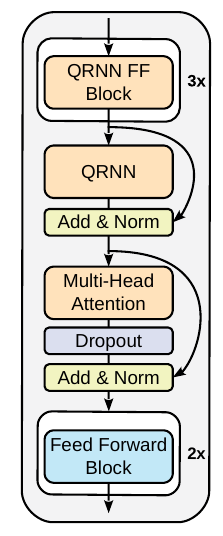}}
  \hfill
  \subfigure[PAR transformer]{\includegraphics[scale=1.0]{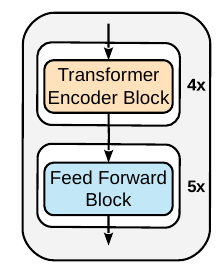}}
  \hfill
  \subfigure[Hybrid Transformer]{\includegraphics[scale=1.0]{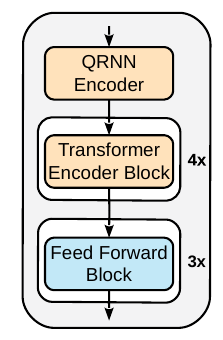}}
  \caption{The "Sequential Network" part of our three language modeling architecture layouts}
  \label{fig:architectures}
\end{figure*}

\paragraph{Three architectures.} Based on these building
blocks, we define two hybrid recurrent-transformer architectures and one transformer baseline, shown in Table~\ref{tab:architectures}. In order of
increasing complexity, they are: (a), the \textbf{Attn-QRNN}, our
adapted version of the SHA-LSTM; the Attn-QRNN model uses almost exclusively recurrent
layers for its sequence processing, with only a single attention layer
near the end of the network \citep{merity2019single} - a QRNN is a quasi-recurrent, parallelizable layer \citep{bradbury2016quasi}. We also use advanced weight dropout (AWD), wich adds recurrence compatible dropout to the QRNN cell itself; (b), the \textbf{PAR (pay attention when required) Transformer}, a transformer variant from prior work which replaces every second multi-headed attention layer with a feed-forward layer, so one attention layer is followed by three feed-forward layers - it has been
found to perform comparably to the standard layout, while speeding up
inference by a factor of 1.5 \citep{mandava2020pay}; and (c), our novel \textbf{Hybrid Transformer}, which consists of two AWD-QRNN blocks \citep{merity2018regularizing}
combined with a PAR Transformer, removing some feed-forward layers for
parameter parity. The basic building blocks are also visualized in Figure~\ref{fig:building_blocks}.
The full architectures can be seen in Figure~\ref{fig:architectures}.

  \begin{table} 
\centering
\begin{tabular}{ll}
\toprule
Model & Architecture\\
\midrule
Attn-QRNN & \texttt{|} + $3\times$(\texttt{q|f}) + (\texttt{qafff})\\
PAR Transformer & \texttt{|} + $4\times$(\texttt{afff}) + $5\times$(\texttt{f})\\
Hybrid Transformer & (\texttt{|q|qf}) + $4\times$(\texttt{afff}) + $3\times$(\texttt{f})\\
\bottomrule
\end{tabular}
\caption{Definition of our three architectures. Letters stand for
  layers: a = relative multi-headed attention, q = AWD-QRNN, f = feed-forward, | = RNN-Dropout
  layer.  
}
\label{tab:architectures}
\end{table}


\section{Experimental Setup}
\label{sec:experiments}

\subsection{Experimental Rationale}

In our experiments, we test the models on two language modeling datasets, enwik8 and WikiText-103 \citep{enwik8, wikitext103}, which both serve as benchmarks for what is essentially a compression evaluation. The goal of language modeling in this context is to measure how well the models predict or compress the data, and these datasets provide a solid foundation for assessing model performance through this lens. Compression is, after all, one of the original evaluation tasks for language models, and we believe it offers a straightforward and interpretable way to gauge their effectiveness without relying on the sometimes ambiguous outcomes of more specialized, modern benchmarks.

We maintain a consistent number of parameters and training time across all models to ensure the results are comparable. Each model is trained in a low-resource setting, with corpora containing roughly 100 million tokens and 41 million model parameters (excluding embeddings). This setup aligns with prior work in the field \citep{dai-etal-2019-transformer,longformer20}. We tune hyperparameters based on performance on the enwik8 dataset -- a more detailed breakdown of this training process can be found in Appendix A. All models are trained for two days on an Nvidia Titan RTX, with the Hybrid Transformer model being 1.7\% slower in terms of training speed per data batch. For the WikiText-103 dataset, we employ weight tying for the input and output embeddings.

\subsection{Datasets}
\label{sec:datasets}

\textbf{WikiText-103} is a corpus of text from the English Wikipedia with a 100 million
word training corpus \citep{wikitext103}. The data only contains article
text without HTML. It has a predefined whole-word
tokenization, and most articles are between 1000 and 7000 tokens
long. We split the text at article boundaries and feed them into
our models as separate sequences bundled into batches.
\textbf{Enwik8} is a character-based 100MB HTML-dump text corpus from
Wikipedia \citep{enwik8}.  It is mainly used to test how well LMs could
hypothetically compress the corpus of 100 million bytes by predicting
the next character given the previous context. As it also uses English Wikipedia
as a source, the make-up of the data is comparable to Wikitext-103, apart from
the HTML code surrounding the articles. We use the common 90/5/5 train/validation/test
split.
As a compression benchmark, enwik8 is meant to be trained as one continuous sequence. Therefore we split up the training set into batch-size sequences.

\subsection{Evaluation Measures}

We report results of a single fixed-seed run.
For enwik8, we evaluated on the bits-per-character (BPC) metric, a
standard metric for language modeling
\cite{DBLP:journals/corr/Graves13}. Equivalent to the logarithm of character-wise perplexity, it is a measure of theoretical text compressability. It is defined as:
$\text{BPC} = -\frac{1}{N} \sum_{i=1}^{N} \log_2 P(character_i)$.
For WikiText-103, the standard evaluation metric is
perplexity (PPL), which measures how well the model output distribution
predicts the ground truth. It is typically defined as:
$\text{PPL} = \exp \left( -\frac{1}{N} \sum_{i=1}^{N} \log P(word_i) \right)$.
We also compute a bits-per-utf-8-byte
compression metric - in the same way as BPC -, to obtain results which are roughly comparable with enwik8. In all cases, lower is better.
For evaluation, we use fixed-length batches of (partial) sequences.

\begin{table*}[h!tb]
\centering
\begin{tabular}{lrllll}
\toprule
\multirow{2}{*}{Model} & \multirow{2}{*}{\#Params} & \multicolumn{2}{c}{Bits-per-Byte} & \multicolumn{2}{c}{BPC}\\ 
&& valid & test & valid & test\\
\midrule
LSTM \citep{merity2019single} & 51M &  &  & 1.312 & 1.33\\
AWD-LSTM \citep{analysis_awdlstm} & 47M  & &  &  & 1.232\\
\midrule
Attn-QRNN (ours) &  40.48M & 1.036 & 1.019 & 1.078 & 1.058\\
PAR Transformer (ours) & 41.18M & 1.026 & 1.009 & 1.068 & 1.047\\
Hybrid Transformer (ours) & 41.44M  & 0.994 & 0.976 & 1.034 & \textbf{1.013}\\
\midrule
Transformer-XL \citep{dai-etal-2019-transformer} & 41M &  &  &  & 1.06\\
Transformer-XL \citep{mandava2020pay} & 41M &  &  &  & 1.10\\
PAR Transformer \citep{mandava2020pay} & 47M &&& & 1.11\\
\midrule
Longformer \citep{longformer20} & 41M &  &  & 1.02 & \textbf{1.00}\\
Adaptive transformer \citep{sukhbaatar-etal-2019-adaptive} & 39M &  &  & 1.04 & 1.02\\
\midrule
SHA-LSTM \citep{merity2019single} & 54M &  &  & 1.096 & 1.068\\
SRU++ Base \citep{lei2021attention} & 108M & &  &  & 0.97\\
SRU++ \citep{lei2021attention} & 42M &  & &  & \textbf{1.022}\\
\bottomrule
\end{tabular}
\caption{Enwik8 results: our models and related work with similar parameter numbers}
\label{tab:enwik8_results}
\end{table*}

\section{Results}
\label{sec:results}

We first discuss the results for enwik8, shown in Table~\ref{tab:enwik8_results}.

All three architectures outperform their respective counterparts from other works: The PAR Transformer from this work achieves a test-set BPC of 1.047 and improves upon the Transformer-XL with same parameter count and a test BPC of 1.06 by 0.013 BPC. It also performs better than the original PAR Transformer, which has 1.11 BPC -- and has more parameters -- although it uses longer attention ranges than the latter and additional RNN-Dropout.

\begin{table*}[tb!]
\centering
\begin{tabular}{lrllll}
\hline
\multirow{2}{*}{Model} & \multirow{2}{*}{\#Params} & \multicolumn{2}{c}{Bits-per-Byte} & \multicolumn{2}{c}{PPL}\\
&& valid & test & valid & test\\
\hline
Attn-QRNN (ours) &  61.1M & 1.054 & 1.078 & 22.27 & 23.37\\
PAR Transformer (ours) & 59.6M & 1.027 & 1.051 & 20.56 & 21.59\\
Hybrid Transformer (ours) &  59.9M  & 1.018 & 1.040 & 20.07 & \textbf{20.91}\\
\hline
Transformer-XL \citep{dai-etal-2019-transformer} & 151M &  &  & 23.1 & 24.0\\
SRU++ Base \citep{lei2021attention} & 148M &  &  &  & 18.3\\
\hline
\end{tabular}
\caption{Wikitext-103 results: our models and related work with similar parameter numbers}\label{tab:wikitext103_results}
\end{table*}

The Hybrid Transformer itself improves with 1.013 BPC up on our PAR transformer by
0.034 BPC, a significant improvement according to Student's t-test (p
$<$ 0.001). While the Attn-QRNN outperforms with 1.058 BPC the SHA-LSTM's 1.068 BPC
\citep{merity2019single}, the latter was trained for only half
the time on a significantly slower GPU. Nevertheless, we find that the
Attn-QRNN reaches almost identical results after only 22 hours of
training. As language models performance scales with data and compute investment \citep{Rae2021Scaling} - which includes in a limited way iterated training over the same data -
we mention if there is a large gap in known compute between models. We do this in the cases where either the architecture is comparable or our model gets outperformed by a model with same parameter count.
The SRU++ variant \cite{lei2021attention}, which is most comparable to
our models, achieves 1.022 BPC and is outperformed by our Hybrid Transformer,
the former requiring 37 GPU hours of training time on Nvidia 2080Ti's.
The Adaptive Transformer by \citet{sukhbaatar-etal-2019-adaptive} likewise underperforms
our Hybrid Transformer with 1.02 BPC.
While the Longformer with 1.00 BPC outperforms the Hybrid Transformer by about 0.01 BPC,
it needs approximately 32 times more compute to train or 16 days on 4 RTX 8000 GPUs \cite{longformer20}.
Overall, the Hybrid Transformer outperforms comparable models, improving significantly on the already strong baseline.

Table~\ref{tab:wikitext103_results} shows results for Wikitext-103. In
the PPL evaluation, all our models outperform the small Transformer-XL
\citep{dai-etal-2019-transformer} with 151 million parameters,
compared to $\approx$ 60 million for our models (including
embeddings).  In contrast, the former uses a much shorter attention
length than our models, 384 vs. 1600 for training and 1600 vs. 2048
for testing. We are not aware of other models that are strictly
comparable.
On this dataset, however, the Hybrid Transformer and PAR Transformer
results are not significantly different in PPL according to a
Kolmogorov-Smirnov test -- they perform essentially on par.

The Bits-per-Byte results match the two experiment-specific metrics,
BPC and PPL, well, and confirm the general order from Hybrid
Transformer (best model) to Attn-QRNN (worst model).


\section{Conclusion and Future Work}
\label{sec:conclusion}

Our paper provides evidence that transformers with fewer
attention layers can benefit significantly from QRNN blocks at the
beginning. 
Compared to other architectures, our hybrid models show
superior performance at the same number of parameters, or the same
performance at fewer parameters, indicating a ``sweet spot'' that can
improve language modeling in terms of both effectiveness and
efficiency.

That being said, a clear limitation of our study is that we have only
considered a relatively low-resource scenario, with datasets of 100M
tokens and models with 40M parameters.  While such a setting is ideal
for examining which models perform best under strict resource
limitations, further work is needed to understand how our hybrid
architectures scale, and to determine if they still provide efficient
alternatives to ``pure'' transformers in higher-resource
regimes. Another relevant aspect is the transferability of our models
to languages that are typologically different from English, which
also requires a closer look at the impact of tokenization.

We see the combination of recurrent layers and attention layers into
carefully-crafted hybrid architectures as an important direction in
developing more efficient models for NLP.  We hope that our work can
help to inspire future work in exploring alternative recurrent units,
examining how best to compose recurrent and attention layers, and
understanding how such hybrid models scale.

\section*{Ethics Statement}

Our work makes an architectural contribution to language modeling,
which is a fundamental task in NLP. On an ethical level, we believe
that two aspects of language modeling are noteworthy:
\begin{itemize}
\item First, the question of fairness: Language models been found to
  be subject to biases of various kinds
  \cite{NEURIPS2020_92650b2e,pmlr-v139-liang21a}. We do not believe
  that our work has a systematic impact on the degree of fairness of
  language models, and thus does not introduce additional risks.
\item Second, the environmental impact of large-scale model training.
  Our work aims at improving the situation in this regard.
\end{itemize}

\bibliography{custom}


\appendix

\section{Hyperparameters \& Training}
\label{sec:ap_hyperparameters}

The hyperparameters of the actual model architectures themselves can be found in
Table~\ref{tab:ap_hyperparams}. \textit{Embedding RNN-Dropout} refers to the
RNN-dropout applied just after the input embedding layer.
For the QRNN layers we always use a convolution size of 2 for the very first QRNN layer and 1 - just $x_t$ - otherwise. Also in case of any \texttt{qf} block, there will be a residual connection going fully around it and a layer-norm being applied to it.

As optimizer for weight updates we use AdamW, an Adam version with improved weight-decay.
We use the default $\beta$ values of Adam \citep{loshchilov2018Adamw, Kingma2015Adam}.
The learning rate schedule is always a one-cycle learning rate, which rises from a start
value to a peak at the first 3rd and then declines again. Similar to a cosine half-period
\citep{Smith2018Onecycle}. 

We use batch-steps of sizes 32 for enwik8 and 16 for
Wikitext-103 with gradient accumulation, as one full batch of 64 doesn't fit into GPU memory.
The learning rate is frozen during those intra batch-steps.
For Wikitext-103 we use an adaptive embedding \cite{baevski2018AdaptiveEmb} and -softmax
\cite{Grave2017AdaptiveSoftmax}, whose splits - embedding size division factor is 4 - can also
be found in Table~\ref{tab:ap_hyperparams}. 

\begin{table*}[h!b]
\centering
\begin{tabular}{llll}
\toprule
Parameter & PAR Transformer & Hybrid Transformer  & Attn-QRNN \\
\midrule
Embedding dim & 512 & 512 & 768\\
FF boom layer dim & 2048 & 2048 & 3072\\
Attention heads & \multicolumn{2}{c}{8} & 12\\
Adaptive Embedding splits & \multicolumn{3}{c}{[20k, 40k, 200k]}\\
\midrule
Dropout & 0.13 & 0.13 & 0.15\\
Embedding RNN-Dropout & 0.3 & 0.3 & 0.35\\
RNN-Dropout & n/a & 0.16 & 0.15\\
RNN Weight-Dropout & n/a & 0.3 & 0.35\\
\midrule
Peak learning rate & \num{4e-4} & \num{4.5e-4} & \num{4.5e-4}\\
Start learning rate & \multicolumn{3}{c}{\num{1e-7}}\\
Final learning rate & \multicolumn{3}{c}{\num{5e-6}}\\
AdamW weight-decay & \num{1e-3} & \num{1e-3} & \num{2e-3}\\
\midrule
Effective Batch-size &  \multicolumn{3}{c}{64}\\
BPTT input length & \multicolumn{3}{c}{512}\\
Enwik8 Train attention length & \multicolumn{2}{c}{768} & 1024\\
Enwik8 Test attention length & \multicolumn{3}{c}{2048}\\
WT103 Train attention length & \multicolumn{2}{c}{384} & 1024\\
WT103 Test attention length & \multicolumn{2}{c}{1600} & 2048\\
\midrule
Enwik8 batch steps & 280k & 275k & 444k\\
Wikitext-103 batch steps & 270.5k & 264.5k & 383.4k\\
\bottomrule
\end{tabular}
\caption{Hyperparameters used for the given models and all datasets}
\label{tab:ap_hyperparams}
\end{table*}


\section{Detailed Results of PG19}
\label{sec:ap_results}

PG-19\footnote{\label{fn:pg19dataset}\url{https://github.com/deepmind/pg19}}
is a book corpus dataset from project Gutenberg.

For PG-19 we use unigram language model sub-word tokenization \citep{kudo-2018-subword}. Compared to BPE encoding this leads to more natural/morphologically correct splits which conform better to actual suffixes or syllables \citep{bostrom-durrett-2020-byte}. We use the SentencePiece library for the Uni-gram encoding.\footnote{\label{fn:sentpiece}\url{https://github.com/google/sentencepiece}} To compare the behaviour of different sub-word lengths we use several different vocabulary sizes -- 1024, 16384 and 32768. We also use an additional byte-level tokenization.
We also train the language models in case of the 16384 token vocabulary with about 2.8\% of the data - mimicing enwik8 character amount - to test the relative performance.

We have chosen byte-level tokenization over character, because PG-19 contains thousands of the latter. The size of 1024 is for testing of a vocabulary of the most frequent (sub-)words. For all vocabulary sizes of 1024 and greater we have only used characters in the 99.95\% percentile. Meta-tokens are included in all mentioned vocabulary sizes.

Looking at the results in Table~\ref{tab:ap_pg19_results}, the Hybrid Transformer gives consistenly the best values, followed by the PAR Transformer. The gap between the models increases for decreasing token lengths.
The test set values are also always much better than those of the validation set. This behaviour also occurs for other works by \cite{Rae2020Compressive} and \cite{roy-etal-2021-efficient}.

Comparing to the latter, it becomes obvious, that the model parameter sizes are far to small for this large dataset. Even a 500 million parameter model only obtains a perplexity of 33.

The Levene tests show that the Hybrid- and PAR Transformer have equal variances for all tokenizations. The p-values are >90\%. The Students t-test confirms the normal distribution for all tokenizations except for the 1024 vocabulary. Although looking at the distribution manually it is very close to a normal one with a few outlier datapoints. We took the logarithm of the data to allieviate this distribution issue.
The Students t-test shows for all tokenizations with p-values < \num{1e-60}, that the Hybrid Transformer is better than the PAR Transformer.

\begin{table*}[tbh!]
\centering
\begin{tabular}{lllllll}
\hline
\multirow{2}{*}{Model} & \multirow{2}{*}{\parbox{2cm}{Vocab Size}} & \multirow{2}{*}{\#Params} & \multicolumn{2}{c}{Bits-per-Byte} & \multicolumn{2}{c}{PPL} \\
&&& valid & test & valid & test\\
\hline
Attn-QRNN (ours) & \multirow{3}{*}{\parbox{2cm}{264 (Bytes)}} & 40.36M & 1.293 & 1.248 & 178.85 & 152.32\\
PAR Transformer (ours) && 41.1M & 1.281 & 1.242 & 170.21 & 148.66\\
Hybrid Transformer (ours) && 41.36M & 1.246 & 1.206 & 147.78 & 128.66\\
\hline
Attn-QRNN (ours) & \multirow{3}{*}{1024} & 40.95M & 1.224 & 1.184 & 135.33 & 117.85\\
PAR Transformer (ours) && 41.49M & 1.213 & 1.171 & 129.79 & 111.65\\
Hybrid Transformer (ours) && 41.75M & 1.191 & 1.149 & 118.55 & 102.90\\
\hline
Attn-QRNN (ours) & \multirow{3}{*}{\parbox{2cm}{16,384 (2,8\% data size)}} & 48.03M & 1.245 & 1.201 & 147.58 & 125.90\\
PAR Transformer (ours) && 49.37M & 1.229 & 1.183 & 138.06 & 117.32\\
Hybrid Transformer (ours) && 49.63M & 1.216 & 1.169 & 131.00 & 110.80\\
\hline
Attn-QRNN (ours) & \multirow{3}{*}{16,384} & 48.03M & 1.181 & 1.143 & 113.84 & 99.78\\
PAR Transformer (ours) && 49.37M & 1.159 & 1.122 & 104.54 & 91.49\\
Hybrid Transformer (ours) && 49.63M & 1.144 & 1.106 & 98.22 & 85.94\\
\hline
PAR Transformer (ours) & \multirow{2}{*}{32,768} & 57.78M & 1.143 & 1.105 & 97.87 & 85.60\\
Hybrid Transformer (ours) && 58.04M & 1.128 & 1.090 & 92.17 & 80.46\\
\hline
Compressive Transf. \citep{Rae2020Compressive} & \multirow{2}{*}{32,000} & \multirow{2}{*}{~523M\footnote{Estimation from their paper, based on our code}} & & & 43.4 & 33.6\\
Transformer-XL \citep{Rae2020Compressive} && & & & 45.5 & 36.3\\
Routing Transformer \citep{roy-etal-2021-efficient} & 98,000 & n/a & & & & 33.2\\
\hline
\end{tabular}
\caption{PG-19 results in comparison to other works}\label{tab:ap_pg19_results}
\end{table*}

\end{document}